# Compositional Stochastic Modeling and Probabilistic Programming

Eric Mjolsness
Department of Computer Science
University of California
Irvine, CA 92697
*emj@uci.edu*

## Abstract

Probabilistic programming is related to a compositional approach to stochastic modeling by switching from discrete to continuous time dynamics. In continuous time, an operator-algebra semantics is available in which processes proceeding in parallel (and possibly interacting) have summed time-evolution operators. From this foundation, algorithms for simulation, inference and model reduction may be systematically derived. The useful consequences are potentially far-reaching in computational science, machine learning and beyond. Hybrid compositional stochastic modeling/probabilistic programming approaches may also be possible.[1]

## 1    Introduction

Programming languages typically have semantics that is compositional, deterministic, and defined in discrete time and space. "Probabilistic programming" (PP) proposes to change the "deterministic" part of this description. If one also changes the "discrete time" part of this description, allowing continuous time which is more natural for most scientific applications, we arrive at a class of computational modeling languages rather than programming languages: compositional stochastic modeling (CSM) languages. Such languages have many attractive features for modeling real-world processes: a close match of computational and physical semantics, a systematic way to derive algorithms for sampling, inference, and model reduction, and relevance to new vistas in biocomputing and abstract mathematics, among others.

Unfortunately the jump between continuous and discrete time seems like a large one – even more so if continuous space is also admitted for modeling purposes. On the other hand conventional computational "engines" are possible for such languages, so perhaps the gap can be closed. What would the consequences be of taking CSM as a model of computation? Obviously such a model would be probabilistic. How can it be related to probabilistic programming, to their mutual enrichment? After a brief review of one approach to CSM, we discuss various points that bear on these questions, and then address the applications that may benefit from CSMs.

## 2    Operator algebra approach to CSM

An example of CSM is the "Plenum" implementation of the "Dynamical Grammars" modeling language [1], which was developed for developmental systems biology but could be applied to many other scientific modeling domains as well [2]. A dynamical grammar (DG) consists of a multiset of reaction-like rewrite rules, each with a computer-algebraically expressed function quantifying the rate (or "propensity") of the corresponding process. Rules can act instantaneously or continuously in time, the latter by way of differential equations. Either way, the effects of a rule may alternatively be defined by (recursively) calling a subgrammar. The semantics of dynamical grammars is defined by mapping each rule to a time-evolution operator that appears in the master equation for the evolution of probabilities on states of the system. (The master equation was proposed independently as a way to provide the semantics for a "small stochastic process algebra" in [3]. However, the present use of operator algebra to express the semantics itself was not part of that proposal.) The algebra of possible time-evolution operators is generated by a set of elementary object creation and annihilation operators, under the operations of operator addition (corresponding to parallel processes), operator multiplication (corresponding to atomic sequential events), and scalar multiplication by rate functions in a defined function space eg. Sobolev Banach space.

---

[1] What follows is an Extended Abstract for the Neural Information Processing Systems (NIPS) Workshop on Probabilistic Programming, 2012.

The resulting semantics is "compositional" in that (a) the operator for a (multi-) set of rules is the (multiplicity-weighted) sum of their operators, so multiset union maps to operator addition; (b) dynamically graph-structured data objects are easily expressed; and (c) subgrammars have suitable semantics for the analog of subroutines and/or macros [1].

The master equation which defines process semantics can be reexpressed in a form, called the time-ordered product expansion (TOPE), which maps the possible interaction histories of system objects to Feynman diagrams [1,4]. This form also permits the derivation of the Gillespie algorithm for stochastic chemical kinetics simulation, and its generalization to DG rewrite rule systems. Between interaction events, continuous-time processes expressed in terms of differential operators become (under TOPE) differential equations (to be solved numerically by conventional discretization) as proved in [4]. Thus the TOPE provides a systematic route for deriving Markov chains for simulation and also maximum-likelihood inference algorithms [2,5,4], from the master equation and operator algebra semantics.

In addition to the *process* semantics defined above, CSM languages may also have an *object* semantics describing the mathematical spaces (topological spaces, measure spaces, geometries, and/or function spaces) that circumscribe its dynamical objects. Within such constraints, we desire type constructors including Cartesian products, disjoint sums, functions (possibly including higher-order functions), and quotients. This topic is discussed at greater length in [6]. These type constructors are naturally expressed in terms of category theory, potentially bringing an entirely different sense of the phrase "algebraic semantics" into play.

Collections of rewrite rules in the form of semi-Thue systems were one of the first formalisms shown to be capable of universal computing. DG rules are each much more powerful than semi-Thue system rules, so DGs are Turing-universal. DGs with unbounded rate functions are capable (in principle) of super-Turing computation, by creating a succession of new objects that speed up according to progressively faster clocks. If the sum of effective clock times (inverse propensity functions) converges, there is an accumulation point through which a conventional computer could not simulate the DG. Accumulation points can themselves accumulate, and so on, creating the image of some countable ordinal as in [7]. But it is fairly natural in DGs to eliminate such problems by enforcing conservation or monotonic decrease of global finite resource parameters as part of the dynamics as well, just as chemical reactions conserve the total amount of each chemical element. Most scientific applications already have such parameters.

## 2      Relationship to probabilistic programming

There is an important mapping of continuous to discrete semantics, by way of the TOPE and the Gillespie algorithm. A discrete-time semantics was defined in [1] which essentially throws away the real-valued event times in a DG without differential equations (a stochastic parameterized grammar or SPG). With this semantics, SPGs in fact comprise an expressive probabilistic programming language. If a different PP language can be reduced (efficiently) to SPGs, it can be reduced (efficiently) to DG's by restoring the missing event times using calls to an Erlang distribution with parameters that must be computed anyway for the discrete-time SPG semantics. In the reverse direction, DGs can be reduced to Turing machines incorporating plentiful calls to random number generators, since simulation engines exist for DGs. The main ingredients in one such engine [2], derivable from TOPE [4] are: variable-binding to define instantiated rule execution probabilities and results, Gillespie stochastic simulation algorithm choice of rule execution, a Rete algorithm like data structure for efficient handling of many rules, and calls to a differential equation solver. If a PPL can implement these items (efficiently) then DGs are reducible to the PPL (efficiently).

Similarities of CSM to PP include the following. *Simulation* algorithms can be derived in a systematic and sometimes automatic way. As in the case of PP, SPGs can have their parameters *inferred* from data about their behavior by a maximum-likelihood algorithm. Such an algorithm has been derived from TOPE and demonstrated on a small gene regulation network problem [5]. Approximate *model reduction* is also possible for some SPGs expressed in Plenum. The reduced model is of a simpler form than a CSM, being much closer to a graphical model. Such model reduction was demonstrated in a synaptic signaling model [8].

Of the foregoing CSM compositionality properties property (a), summation of rule time-evolution operators, is compromised by the move from continuous to discrete time. This fact may result in improved technical tractability in the continuous-time case, as well as greater congruence with real-world continuous-time and asynchronously parallel applications. Hybrid discrete-time/continuous-time DG's are also possible, provided that they are well separated by the encapsulation mechanism provided by subgrammar (not macro) calls.

Similar work in the direction of complex continuous-time stochastic models with time-changing numbers of random variables is not yet an overpopulated category. Continuous-Time Bayes Nets (CTBNs) [9] have a time-invariant number of discrete random variables and therefore are described by a fixed-dimension intensity matrix. CTBNs have been generalized to certain nonexponential delay distributions $P(\Delta t)$ eg. Erlang [10]. Coalescent theory in genetics is also continuous-time and stochastic, but relies on highly domain-specific assumptions. More general continuous-time Markov process frameworks still have a fixed and usually finite or at most countably infinite state space. Stochastic pi-calculus [3] has time-changing variable number and semantic similarities discussed above, but no rate function spaces, submodels, etc. CTPPL [11] is another generalization of CTBNs to arbitrary expressions for delay distributions $P(\Delta t)$ and also time-changing data structures. Such delay distributions may be handled in DG by equivalent time-varying rate functions $\rho(\Delta t) = P(\Delta t)/[1-\int_0^{\Delta t} P(\tau)d\tau]$ ; if $P(\Delta t)$ is Erlang, for example, then $\rho(\Delta t;n,\lambda) = \lambda^n t^{n-1} e^{-\lambda t}/\Gamma(n,t\lambda)$ where $\Gamma$ is the incomplete gamma function and of course $\Delta t$ is nonnegative. None of these frameworks handle continuous variables, nor their possibly continuous evolution in time according to differential (or stochastic differential) equations, as does the Dynamical Grammar framework.

## 3   Relevant application domains

The kind of compositional stochastic modeling language discussed here has potentially broad applicability due to the additive compositionality of its semantics (parallel processes have summed time-evolution operators), which along with its continuous-time model makes for a close match of computational semantics and dynamics in scientific domains (physical, electronic, biological, social, etc.). Algorithms for simulation, inference, and model reduction follow. Biological applications have been demonstrated with models of gene regulation, molecular complexes in synapses, and tissues comprising plant and animal stem cell niches with cell division and diffusible growth factors [2,12]. A schematic example of one such stem cell niche application is the dynamical grammar of Figure 1, explained more fully in [6].

Other SPG-like "rule-based" biochemical modeling languages have been applied to a variety of biological modeling problems at the cellular and molecular levels [13,14]. We may expect progressively more advanced computational biology and biocomputing applications. Varieties of modeling encompassed include stochastic chemical kinetics, dynamical systems given by ordinary differential equations, agent-based models, stochastic string, tree and graph rewrite rules, spatially continuous models (so far only the diffusion equation and mass-spring elastodynamics), and perhaps most importantly, hybrids of all of these types. Thus, CSML's may function as high-level domain-specific languages for computational science.

```
grammar Epithelium {
    /* cell replication with or without differentiation : */
```

$$\text{cell}[\chi : \mathbb{N}, \mathbf{x} : \mathbb{R}^d, V : \mathbb{R}, \phi : \mathbb{R}] \longrightarrow \text{cell}[\chi + \Delta\chi_1, \mathbf{x} + \Delta\mathbf{x}, V/2, \phi],$$
$$\text{cell}[\chi + \Delta\chi_2, x - \Delta\mathbf{x}, V/2, \phi]$$
$$\textbf{with} \quad \hat{\rho}(V) P(\Delta\chi_1 : \mathbb{N}, \Delta\chi_2 : \mathbb{N}|\chi, \phi) \mathcal{N}(\Delta\mathbf{x} : \mathbb{R}^d; cV^{1/d})$$
$$\times \Theta(\chi < \chi_{\max}) \Theta(\Delta\chi_1 \in \{0,1\}) \Theta(\Delta\chi_2 \in \{0,1\})$$

/* cell death : */

$$\text{cell}[\chi_{\max}, \mathbf{x}, V, \phi] \longrightarrow \varnothing \quad \textbf{with} \quad \gamma$$

/* cell growth, dependent on cell type $\chi$ : */

$$\text{cell}[\chi, \mathbf{x}, V, \phi] \longrightarrow \text{cell}[\chi, \mathbf{x}, V + dV, \phi]$$
$$\textbf{solving} \quad \{\tfrac{dV}{dt} = k\Theta(\chi < \chi_{\max}) + k_\varsigma(V)\Theta(\chi = \chi_{\max})\}$$

/* symmetric cell-to-cell diffusion of growth inhibition signal $\phi$ : */

$$\text{cell}[\chi_1, \mathbf{x}_1, V_1, \phi_1], \text{cell}[\chi_{\max}, \mathbf{x}_2, V_2, \phi_2]$$
$$\longrightarrow \text{cell}[\chi_1, x_1, V_1, \phi_1 + d\phi_1], \text{cell}[\chi_{\max}, \mathbf{x}_2, V_2, \phi_2]$$
$$\textbf{solving} \quad \{\tfrac{d\phi_1}{dt} = D(\|\mathbf{x}_1 - x_2\|, (V_1 V_2)^{1/(2d)}) \quad (\phi_2 - \phi_1)\}$$

/* signal production, dependent on cell type $\chi$, and degradation : */

$$\text{cell}[\chi, \mathbf{x}, V, \phi] \longrightarrow \text{cell}[\chi, \mathbf{x}, V, \phi + d\phi]$$
$$\textbf{solving} \quad \{\tfrac{d\phi}{dt} = k'\Theta(\chi = \chi_{\max}) - \lambda\phi\}$$

}

*Figure 1. Schematic example (from [6]) of a dynamical grammar for a stem cell niche model. This model incorporates both discrete events (occurring at a definite instant in continuous time) such as cell division "**with**" a time-varying propensity function, and also extended-duration processes modeled by "**solving**" one or more differential equations. Each object has both discrete-valued parameters (cell type) and continuous-valued parameters (cell position, size, and internal concentration of growth hormone). Model simplified from the olfactory epithelium model of [2].*

The advantages claimed here for CSML's over PP languages for scientific modeling are not advantages of in-principle generality, except in the potentially problematic sense of super-Turing computation, but rather of *perspicuity* in concisely expressing real-world dynamical models, due to the closer of match of continuous-time dynamics to the intended application domains. Thus, CSML's may function as a "higher-level" language in many modeling domains. On the other hand occasionally (as in Poincare maps) discrete-time models of continuous-time systems are also revealing. Therefore one may predict that the translation path between CSML's and PPL's will become well-trodden.

# 4     Further Discussion

One criterion for useful generality in a formal modeling language is that the language should be closed under "expected" kinds of model reduction. Model reductions in physics often change from deterministic to stochastic dynamics or back, from discrete to continuous variables or the reverse, and so on – so we have proposed a framework that encompasses all these variations including hybrids. Oddly, model reductions often proceed by taking *infinite limits*. Approximation of integer molecule numbers by continuous-valued concentrations is one example. Another is spatially continuous PDE models of elastic materials that are actually composed of discrete atoms and molecules. Unfortunately the specter of super-Turing computing will be raised whenever an infinite limit is taken which, in applied mathematics, is quite often; therefore a general modeling framework must permit this kind of trouble. In practice domain-dependent restrictions such as resource constraints, spatial frequency limits, Sobolev norms, and other schemes of regularization are used to control these potentially problematic limits. By working with computer algebra representations it should be possible to express and analyze these kinds of necessary conditions, as indeed has already been achieved for example in PDE packages that support eg. weak forms [15].

Under further development one may expect the following *inherent* capabilities to added to the practical repertoire of CSM languages with operator-algebra semantics: hybrid discrete-time (PP) and continuous-time CSM languages; increasingly general PDEs including dynamic boundaries; application to asynchronous parallel computing; and novel learning and evolution methods which amplify small signals out of large noise backgrounds to create increasingly complex learned algorithms. More speculatively one might also attempt to develop: CSM meta-programming using meta-rules (particular meta-rules in Dynamical Grammars are demonstrated in [2]); the use of Hausdorff topological spaces in the computational theory of types (where non-Hausdorff spaces are usually employed [16]); and perhaps eventually, application of operator algebra semantics to mathematical foundations by which a great variety of mathematical objects acquire dynamics and become more widely understandable – thus advancing the computational reification of mathematics. In all of these projects, compositional stochasticity with operator algebra semantics is an essential ingredient.

*Acknowledgements*
Research was supported by NIH grants R01 GM086883 and P50 GM76516 to UC Irvine, and by a Moore Distinguished Scholar visiting appointment at the California Institute of Technology. I also wish to acknowledge the hospitality, travel support, and research environments provided by the Center for Nonlinear Studies (CNLS) at the Los Alamos National Laboratory, the Sainsbury Laboratory Cambridge University, and the Pauli Center for Theoretical Studies at ETH Zürich and the University of Zürich.